\documentclass[runningheads]{llncs}
\usepackage{graphicx}
\newtheorem{assumption}{Assumption}
\usepackage{amsmath}
\usepackage{algorithm}
\usepackage{algorithmic}
\usepackage{multirow}
\usepackage{booktabs}
\usepackage{wrapfig}

\usepackage{graphicx}
\usepackage{float}

\newfloat{figtab}{htb}{fgtb}
\makeatletter
\newcommand\figcaption{\def\@captype{figure}\caption}
\newcommand\tabcaption{\def\@captype{table}\caption}
\makeatother

\makeatletter
\newcommand*{\indep}{%
  \mathbin{%
    \mathpalette{\@indep}{}%
  }%
}
\newcommand*{\nindep}{%
  \mathbin{
    \mathpalette{\@indep}{\not}
  }%
}
\newcommand*{\@indep}[2]{%
  \sbox0{$#1\perp\m@th$}
  \sbox2{$#1=$}
  \sbox4{$#1\vcenter{}$}
  \rlap{\copy0}
  \dimen@=\dimexpr\ht2-\ht4-.2pt\relax
  \kern\dimen@
  {#2}%
  \kern\dimen@
  \copy0 %
}

\begin{document}
\title{Assessing Classifier Fairness with Collider Bias\thanks{Supported by Australian Research Council (DP200101210) and Natural Sciences and Engineering Research Council of Canada.}}
%
%
\author{Zhenlong Xu$^\dagger$\inst{1} \and
	Ziqi Xu$^\dagger$\inst{1}\orcidID{0000-0003-1748-5801} \and
	Jixue Liu\inst{1}\orcidID{0000-0002-0794-0404} \and
	Debo Cheng\inst{1}\orcidID{0000-0002-0383-1462} \and
	Jiuyong Li\inst{1}\orcidID{0000-0002-9023-1878} \and\\
	Lin Liu\inst{1}\orcidID{0000-0003-2843-5738} \and
	Ke Wang\inst{2}}
\authorrunning{Z. Xu et al.}
%
\institute{University of South Australia, Adelaide, Australia 			  
	\email{\{Zhenlong.Xu,Ziqi.Xu,Debo.Cheng\}@mymail.unisa.edu.au,}\\
	\email{\{Jixue.Liu,Jiuyong.Li,Lin.Liu\}@unisa.edu.au}\and
	Simon Fraser University, Burnaby, Canada\\ \email{wangk@sfu.ca}}
\maketitle

\def\thefootnote{$\dagger$}\footnotetext{Ziqi Xu and Zhenlong Xu contributed equally to this paper.}

\begin{abstract}
The increasing application of machine learning techniques in everyday decision-making processes has brought concerns about the fairness of algorithmic decision-making. This paper concerns the problem of collider bias which produces spurious associations in fairness assessment and develops theorems to guide fairness assessment avoiding the collider bias. We consider a real-world application of auditing a trained classifier by an audit agency. We propose an unbiased assessment algorithm by utilising the developed theorems to reduce collider biases in the assessment. Experiments and simulations show the proposed algorithm reduces collider biases significantly in the assessment and is promising in auditing trained classifiers.

\end{abstract}

\section{Introduction}
\label{sec:introduction}

\begin{wrapfigure}{r}{0.4\textwidth}
    \vspace{-1.1cm}
    \begin{center}
        \includegraphics[width=\linewidth]{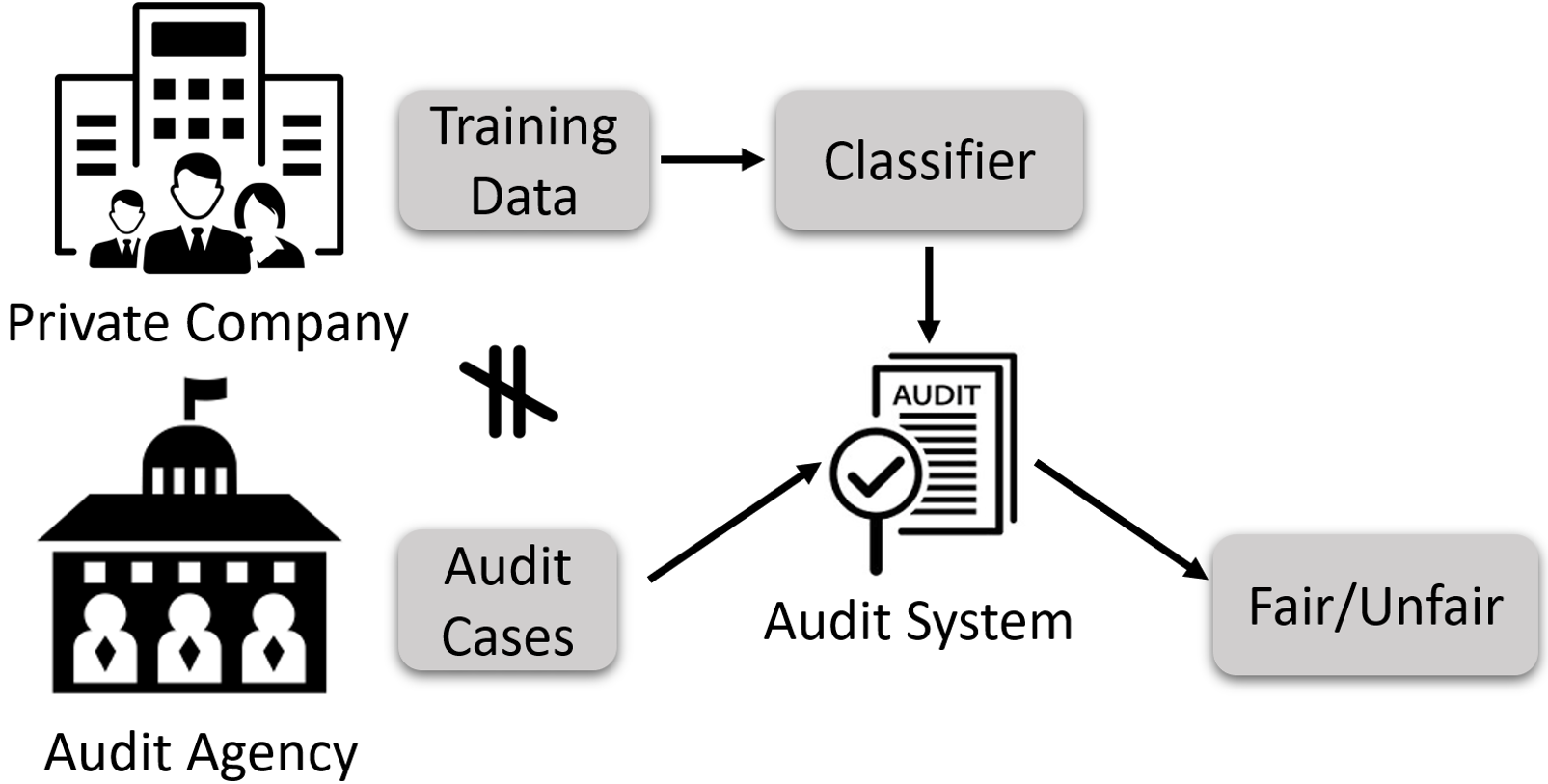}
    \end{center}
    \vspace{-0.5cm}
    \caption{The process of audit.}
    \label{pic:Audit_process}
    \vspace{-0.8cm}
\end{wrapfigure}

There are increasing concerns over the fairness of decision making algorithms with the wide use of machine learning in various applications, such as job hiring, credit scoring and home loan since discrimination can be inadvertently introduced into machine learning models. To prevent unfairness in a model from spreading in society, audit techniques are needed for the independent authority to audit machine learning models. Figure~\ref{pic:Audit_process} shows an audit process. An audit agency accesses a model of a company and has its own audit cases for assessing the fairness of the model. The audit agency does not have access to the training data set but has the regulatory policy. In this paper, we use a causal graph to represent the regulatory policy. The company may use additional variables that are not specified in the regulatory policy to build its models to improve prediction accuracy.

Situation test has been used in the U.S. to detect discrimination in recruitment~\cite{bendick2007situation}, which is a controlled experiment approach for analysing employers' decisions on job applicants' characteristics, as illustrated with the following examples. Pairs of research assistants are sent to apply for the same job, and each pair of the pretended applicants have the same qualifications and experience related to the job but have different values for their protected variable, such as male/female or young/old. Discrimination is detected if the favourable decisions are unequal between groups with different protected values.     

The above described situation test can be simulated in an audit process, and we call it Naive Situation Test (NST) in this paper. We feed two inputs representing two individuals whose variable values are identical except their protected values to a machine learning model. If the model provides different decisions, NST will detect the model as discriminatory.    

\begin{wraptable}{r}{0.39\textwidth}
	\vspace{-1.1cm}
	\caption{An example of incorrect detection by NST on a classifier.}
	\label{tab:biasedNSTdetection}
	\vspace{-0.2cm}
	\begin{center}
		{\scriptsize \begin{tabular}{cccc}
				\specialrule{0.1em}{0pt}{0.3pt}
				Race & Edu & Sub & Predicted.Sal \\ 
				\specialrule{0.05em}{-1pt}{0.3pt}
				white & high & A & \textgreater{}50k \\
				black & high & A & \textgreater{}50k \\
				white & high & A & \textgreater{}50k \\
				black & high & B & $\le$50k \\
				black & high & B & $\le$50k \\
				white & high & B & $\le$50k \\ 
				\specialrule{0.1em}{-1pt}{0.3pt}
			\end{tabular}
			
			\begin{tabular}{c}
				\vspace{-0.2cm}
				\\ NST $\Rightarrow$ ``fair" \\
				\specialrule{0.1em}{0pt}{0.3pt}
				$f$(\textbf{white}, high, A)$=$$f$(\textbf{black}, high, A) \\
				\specialrule{0.05em}{0pt}{0.5pt}
				$f$(\textbf{white}, high, B)$=$$f$(\textbf{black}, high, B) \\ 
				\specialrule{0.1em}{-1pt}{0.3pt}
		\end{tabular}}
	\end{center}
	\vspace{-1.1cm}
\end{wraptable}


NST may produce an incorrect detection. We use the following example to show this. Consider a classifier $f()$ used by a company to determine employees' salaries as $salary=f(race, education, suburb)$. Some predicted outcomes by the model are shown in Table~\ref{tab:biasedNSTdetection}. Based on NST, the black people are not discriminated against since with the same education and suburb, both white and black people are predicted to have the same salary. However, Suburb is an irrelevant variable for determining the Salary. Without considering the Suburb, with the same level of Education, 2/3 white people receive a salary higher than 50K while only 1/3 black people receive a salary of 50K or higher. Hence, black people are discriminated against by the model. 

\begin{wrapfigure}{r}{0.35\textwidth}
	\vspace{-1.1cm}
	\begin{center}
		\includegraphics[width=0.95\linewidth]{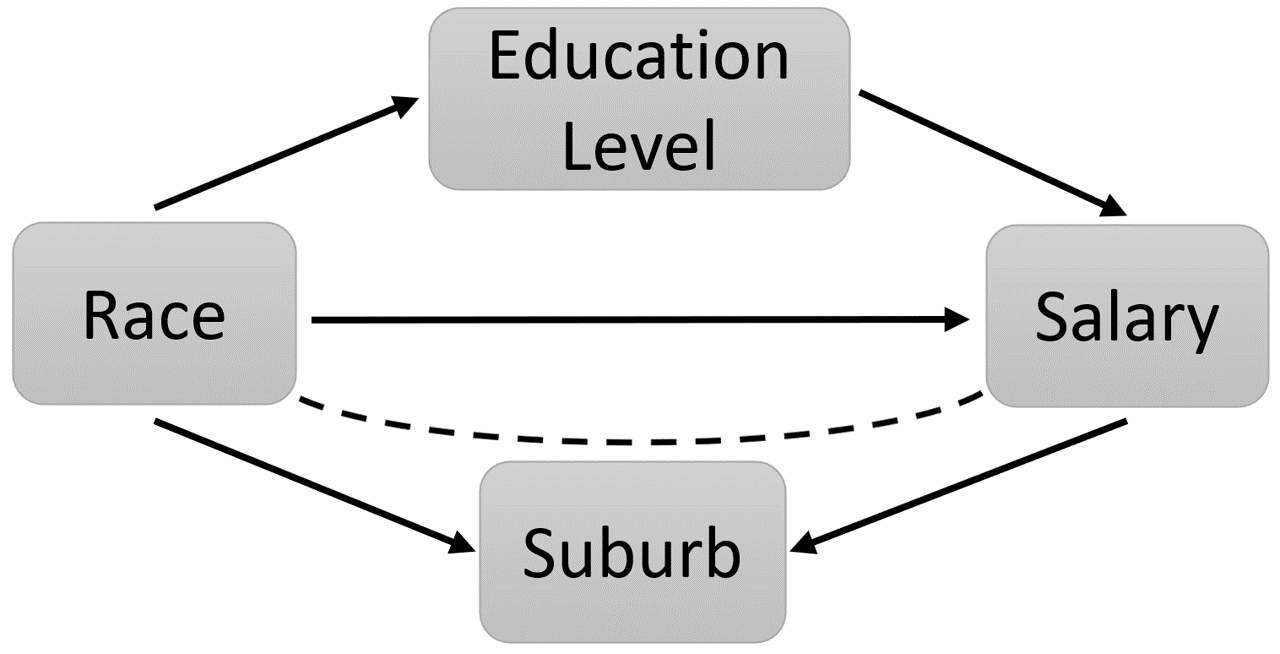}
	\end{center}
	\vspace{-0.6cm}
	\caption{The causal graph for the above example.}
	\label{pic:Causal_graph}
	\vspace{-0.6cm}
\end{wrapfigure}


The incorrect detection by NST is caused by collider bias. We use a causal graph, formally defined in Section~2, to explain the collider bias. Causal relationships of variables in the above example are shown using the causal graph in Figure~\ref{pic:Causal_graph} where a directed edge represents a causal relationship. The suburb is a collider since two edges ``collide" at it. Conditioning on a collider, an association is formed between the two variables but it is spurious~\cite{ConditioningOnCollider}. In the example, the spurious association cancels the association due to the causal relationship between Race and Salary and hides the true discrimination. Collider bias is related to the selection bias~\cite{hernan2020causal}. In a classifier, conditioning on a variable is equivalent to selecting sub-populations using the values of the variable. If the variable is a collider, a selection bias in each sub-population is resulted. We call this bias collider bias in this paper.

There is no existing alternative method to NST to audit classifiers. Most methods, to be reviewed in the related work, need to access the training data set and suffer from collider bias. A causal-based situation test (CST) \cite{zhang2016situation} does not suffer from collider bias, but needs to access the training data set too. The audit cases used by an audit agency are only a small number of individual cases which do not represent the population. Collecting a representative sample of the population needs a significant resource. Therefore, a data-based audit  method is not applicable. We make the following contributions in this paper. 

\begin{itemize}
\item We study collider bias in fairness assessment and present theorems to avoid collider bias. Our theoretical results give a principled guidance on which variables can be used for fairness assessment and also for building fair classifiers.
	
\item We investigate the problem of auditing machine learning models and propose an Unbiased Situation Test (UST) algorithm for auditing without accessing training data or an unbiased sample of the population. Experiments show that UST can effectively reduce collider bias.
\end{itemize}

\section{Background}
\label{sec:background}
We present the necessary background of causal inference. We use upper case letters to represent variables and bold-faced upper case letters to denote sets of variables. The values of variables are represented using lower case letters.

Let $\mathcal{G}=(\mathbf{V},\mathbf{E})$ be a graph, where $\mathbf{V}=\{V_{1},\dots, V_{p}\}$ is the set of nodes and $\mathbf{E}$ is the set of edges between the nodes, i.e. $\mathbf{E}\subseteq \mathbf{V}\times \mathbf{V}$. A path $\pi$ is a sequence of distinct nodes such that every pair of successive nodes are adjacent in $\mathcal{G}$. A path $\pi$  is a directed path if all edges along the path are directed edges. A path between $(V_i,  V_j)$ is a backdoor path with respect to $X_i$ if it has an arrow into $V_i$. Given a path $\pi$, $V_{k}$ is a collider node on $\pi$ if there are two edges incident like $V_{i}\rightarrow V_{k} \leftarrow V_{j}$. In $\mathcal{G}$, if there exists $V_i\rightarrow V_j$, $V_i$ is a parent of $V_j$ and we use $Pa(V_j)$ to denote the set of all parents of $V_j$. In a directed path $\pi$, $V_i$ is an ancestor of $V_j$ and $V_j$ is a descendant of $V_i$ if all arrows point to $V_j$. 

A DAG (Directed Acyclic Graph) is a directed graph without directed cycles. With the following assumptions, a DAG links to a distribution. 

    \begin{definition} [Markov condition~\cite{pearl2009causality}]
    \label{Markovcondition}
        Given a DAG $\mathcal{G}=(\mathbf{V}, \mathbf{E})$ and $P(\mathbf{V})$, the joint probability distribution of $\mathbf{V}$, $\mathcal{G}$ satisfies the Markov condition if for $\forall V_i \in \mathbf{V}$, $V_i$ is  probabilistically independent of all non-descendants of $V_i$, given the parents of $V_i$.
    \end{definition}

When the Markov condition holds, $P(\mathbf{V})$ can be factorized into: $P(\mathbf{V}) = \prod_i P(V_i \mid Pa(V_i))$.  

    \begin{definition}[Faithfulness~\cite{spirtes2000causation}]
    \label{Faithfulness}
        A DAG $\mathcal{G}=(\mathbf{V}, \mathbf{E})$ is faithful to $P(\mathbf{V})$ iff every independence presenting in $P(\mathbf{V})$  is entailed by $\mathcal{G}$ which fulfills the Markov condition. A distribution $P(\mathbf{V})$ is faithful to a DAG $\mathcal{G}$ iff there exists DAG $\mathcal{G}$ which is faithful to $P(\mathbf{V})$.
    \end{definition}
    
With the above two assumptions, we can read the independencies between variables in $P(V)$ from a DAG using the following defined $d$-separation criterion. 
    
    \begin{definition}[$d$-separation~\cite{pearl2009causality}]
	\label{d-separation}
	A path $\pi$ in a DAG is said to be $d$-separated (or blocked) by a set of nodes $\mathbf{Z}$ iff (1) $\pi$ contains a chain $V_i\rightarrow V_k\rightarrow V_j$ and a fork $V_i\leftarrow V_k\rightarrow V_j$ node such that the middle node $V_k$ is in $\mathbf{Z}$, or (2) $\pi$ contains a collider $V_k$ such that $V_k$ is not in $\mathbf{Z}$ and no descendant of $V_k$ is in $\mathbf{Z}$.
	\end{definition}

To conduct causal inference with DAGs, we make the following assumptions.

\begin{definition}[Causal sufficiency~\cite{spirtes2000causation}]
	\label{Causalsufficiency}
	A data set satisfies causal sufficiency if for every pair of variables $(V_i, V_j)$ in $\mathbf{V}$, all their common causes are also in $\mathbf{V}$.
\end{definition}
    
With a DAG, if we interpret a node's parent as its direct cause, the DAG is known as a causal DAG. We can learn a causal DAG from data when the assumptions of causal sufficiency, faithfulness and Markov condition are satisfied.

An intervention, which forces a variable to take a value, can be represented by a \emph{do} operator. For example, $do(X=1)$ means $X$ is intervened to take value 1. $P(y \mid do (X=1))$ is an interventional probability. Let us understand $do$ in an ideal experiment. 

\begin{definition}[Direct effect~\cite{pearl2009causality}]
	\label{def:ATC}
	The direct effect of $X$ on $Y$ is $P(y \mid do (X=x), do (\mathbf{V}_{\backslash XY} = \mathbf{v}))$ where $\mathbf{V}_{\backslash XY}$ means all other variables except $X$ and $Y$.
\end{definition}

In order to study the relationship between $X$ on $Y$, all other variable are controlled in the ideal experiment. To infer interventional probabilities (by reducing them to normal conditional probabilities) with a causal DAG, the following set of rules of $do$-calculus~\cite{pearl2009causality} are necessary. They are used in the proofs of our theorems. 

For a DAG $\mathcal{G}$ and a subset of nodes $\mathbf{X}$ in $\mathcal{G}$, $\mathcal{G}_{\overline{\mathbf{X}}}$ represents the DAG obtained by deleting from $\mathcal{G}$ all arrows pointing to nodes in $\mathbf{X}$, and $\mathcal{G}_{\underline{\mathbf{X}}}$ denotes the DAG by deleting from $\mathcal{G}$ all arrows emitting from nodes in $\mathbf{X}$. 


    \begin{theorem}[Rules of $do$-Calculus~\cite{pearl2009causality}]
    \label{pro:001}
    Let $\mathbf{X}, \mathbf{Y}, \mathbf{Z}, \mathbf{W}$ be arbitrary disjoint sets of variables in a causal DAG $\mathcal{G}$. The following rules hold, where $\mathbf{x}, \mathbf{y}, \mathbf{z}, \mathbf{w}$ are the shorthands of $\mathbf{X}=\mathbf{x}, \mathbf{Y}=\mathbf{y}, \mathbf{Z}=\mathbf{z}$ and $\mathbf{W}=\mathbf{w}$ respectively.
    
    Rule 1. (Insertion/deletion of observations): 

    $P(\mathbf{y} \mid do(\mathbf{x}),\mathbf{z},\mathbf{w}) = P(\mathbf{y} \mid do(\mathbf{x}),\mathbf{w})$, if $(\mathbf{Y} \indep \mathbf{Z} \mid \mathbf{X},\mathbf{W})$ in $\mathcal{G}_{\overline{\mathbf{X}}}$.
    
    Rule 2. (Action/observation exchange):

    $P(\mathbf{y} \mid do(\mathbf{x}), do(\mathbf{z}), \mathbf{w}) = P(\mathbf{y} \mid do(\mathbf{x}),\mathbf{z}, \mathbf{w})$, if $(\mathbf{Y}\indep \mathbf{Z} \mid \mathbf{X},\mathbf{W})$ in $\mathcal{G}_{\overline{\mathbf{X}}\underline{\mathbf{Z}}}$.

    Rule 3. (Insertion/deletion of actions):

    $P(\mathbf{y} \mid do(\mathbf{x}),do(\mathbf{z}), \mathbf{w}) = P(\mathbf{y} \mid do(\mathbf{x}), \mathbf{w})$, if $(\mathbf{Y}\indep \mathbf{Z} \mid \mathbf{X}, \mathbf{W})$ in $\mathcal{G}_{\overline{\mathbf{X}\mathbf{Z}(\mathbf{W})}}$, where $\mathbf{Z}(\mathbf{W})$ is the nodes in $\mathbf{Z}$ that are not ancestors of any node in $\mathbf{W}$ in $\mathcal{G}_{\overline{\mathbf{X}}}$.
	\end{theorem}

\section{Problem Definition}
\label{sec:prob}
A classifier (prediction model) has been built by a company/organisation from a training data set which contains a binary protected variable $A$, a binary decision outcome $Y$, and a set of relevant variables of $Y$, $\mathbf{X}$, since variables independent of $Y$ are not used for predicting $Y$.  An agency wants to audit the model using some cases. We make the following assumptions about the audit. 


\begin{assumption}
	\label{assumption1}
    \begin{enumerate}
    \item The regulatory policy has specified the causal relationships among the factors and $Y$, and uses a causal DAG to indicate. The factors are ancestral variables of $Y$ including all direct causes of $Y$. 
    \item The audit agency has no access to the model training data or an unbiased sample of the population. The agency however has access to the distributional statistics from some sources, such as government census data. 
    \item The company or organisation has used all the legitimate factors to comply with the regulatory policy. However, some other variables are also used by the model to enhance the prediction performance. 
    \end{enumerate}	   
\end{assumption}

In the theorem development, we assume that there is a DAG that is consistent with the regulatory policy. In the algorithm, we do not need the complete DAG, but ancestral variables of $Y$ and colliders in the descendant nodes of $Y$. 

We first define the criterion for auditing. We use Controlled Direct Effect (CDE) \cite{PearlMackenzie18} to measure fairness. CDE is extended from conditioning on \emph{do}(prote\\cted) to conditioning on \emph{do}(all other variables) as in Definition~\ref{def:ATC} to simulate an ideal experiment. The alternative definitions are path specific causal effect~\cite{chiappa2019path,wu2019pc,zhang2018causal} and counterfactual fairness~\cite{kusner2017counterfactual}, we will discuss why the alternatives have not been used after Definition 5. 

The protected variables in this paper include redline variables, which are the descendants of protected variables. The redline variables are recognised as a proxy of protected variables and may cause some discrimination~\cite{kilbertus2017avoiding}. Some companies or organisations build the models under the concept of fairness through awareness~\cite{dwork2012fairness}, which means the classifier functions may not use the protected variables as input. In this case, the redline variables will be considered as the protected variables. 

\begin{definition}[Fairness score]
	\label{def:DS}
	  Given a causal DAG ${\mathcal{G}}$ representing the regulatory policy, $A$, $\mathbf{X}$, and $Y$ as described above. The fairness score is of an individual (or a subgroup) $\mathbf{X = x_i}$ is defined by Controlled Direct Effect, $CDE (\mathbf{x_i}) = P(y \mid do(A=1), do(\mathbf{X}=\mathbf{x}_{i})) - P(y \mid do(A=0), do(\mathbf{X}=\mathbf{x}_{i}))$, where $y$ denotes $Y=1$.
\end{definition}

The rationale of the above definition is that we conduct a controlled experiment by intervening the protected variable, and controlling all other variables to $\mathbf{x}_{i}$. The decision for $\mathbf{x}_{i}$ is fair if the intervention does not change the outcome. 

Unlike previous works~\cite{dwork2012fairness,feldman2015certifying,hardt2016equality,luong2011k}, our definition of fairness score is based on the CDE which uses intervention. Thus the spurious association between $A$ and $Y$ caused by conditioning on colliders will be avoided. We do not use counterfactual fairness \cite{kusner2017counterfactual} in our fairness definition since it needs stronger assumptions and poses a practical challenge. To estimate counterfactual outcomes, there is a need for knowing the full causal model and latent background knowledge. Both are not available in our problem setting.  Some other definitions~\cite{chiappa2019path,wu2019pc,zhang2018causal} make use of path specific causal effect. Their solutions also need counterfactual reasoning and they do not fit our problem setting. 

\begin{definition}[Problem definition]
	\label{def:PS}
	Given ${\mathcal{G}}$, $A$ and $\mathbf{X}$ as described above, and classier $\hat{Y} = f(A, \mathbf{X})$. The audit is to determine if a prediction on an individual ($\mathbf{X = x_i}$) is fair, i.e. $|CDE (\mathbf{x_i})| < \tau$ where $\tau$ is a threshold determined by the regulatory policy and $Y$ in $CDE (\mathbf{x_i})$ is replaced by $\hat{Y}$. 
\end{definition}

\section{Estimating CDE}
\label{sec:method}
For the sake of fairness audit, the protected variable $A$ is assumed to be a parent node of $Y$ so we can use CDE for the audit. The results in this section are true in general, not just for auditing classifiers.  


\begin{theorem}
    \label{theo:main}
    DAG $\mathcal{G}$ contains variables $A$ and $Y$, and variable set $\mathbf{X}$ where $(A \cup Y) \cap \mathbf{X} = \emptyset$. The causal sufficiency is satisfied. $P(y \mid do(A=a), do(\mathbf{X=x})) = P(y \mid A=a,  Pa'(Y) = \mathbf{pa})$ where $Pa'(Y)$ is the set of all parents of $Y$ in $\mathcal{G}$ excluding $A$.
\end{theorem}

\begin{proof}
    Firstly, let $\mathbf{X} = \{\mathbf{C}\cup \mathbf{Q}\}$ where $\mathbf{C}$ contains descendant nodes of $Y$, and $\mathbf{Q}$ contains non-descent nodes of $Y$. We have $P(y \mid do(A=a), do(\mathbf{C=c}), do(\mathbf{Q=q}))\\=P(y \mid do(A=a),do(\mathbf{Q=q}))$. This is achieved by repeatedly using Rule 3 of Theorem~\ref{pro:001}. We show this by an example where $C \in \mathbf{C}$, $P(y \mid do(A=a), do(C=c), do(\mathbf{Q=q})) = P(y \mid do(A=a), do(\mathbf{Q=q}))$ because $Y \indep C$ in DAG $\mathcal{G}_{\overline{A}, \overline{C}}$ where the incoming edges to $A$ and to $C$ have been removed.
    
    Secondly, we consider $P(y \mid do(A=a), do(\mathbf{Q=q}))$ only. Based on the Markov condition~\ref{Markovcondition}, $Y$ is independent of all its non-descendant nodes given its parents. Therefore, $P(y \mid do(A=a), do(\mathbf{Q=q})) = P(y \mid do(A=a), do(Pa'(Y) = \mathbf{pa}))$.
    
    Thirdly, we will prove $P(y|do(A=a),do(Pa'(Y)=\mathbf{pa}))=P(y \mid A=a,  Pa'(Y)=\mathbf{pa})$. This can be achieved by repeatedly applying Rule 2 of Theorem~\ref{pro:001}.
    
    Let $Pa(Y) = \{A, X_1, X_2, \ldots, X_k\}$.
    \begin{equation*}
        \begin{aligned}
            &P(y \mid do(A=a),do(X_1=x_1),do(X_2=x_2),\ldots, do(X_k=x_k))  \\
            &=P(y \mid A=a,do(X_1=x_1)do(X_2=x_2),\ldots, do(X_k=x_k)) \\
            &Since~Y \indep A | X_1, X_2,\dots, X_k~in~\mathcal{G}_{\overline{X_1},\overline{X_2},\ldots,\overline{X_k},\underline{A}} \\
            &=P(y \mid A=a,X_1=x_1,do(X_2=x_2),\ldots,do(X_k=x_k)) \\
            &Since~Y \indep X_1 | A, X_2,\dots, X_k)~in~\mathcal{G}_{ \overline{X_2},\ldots, \overline{X_k},\underline{X_1}} \\
            &Repeat~k-1~times \\
            &=P(y \mid A=a,X_1=x_1, X_2=x_2,\ldots,X_k=x_k) \\
            &=P(y \mid A=a,Pa'(Y)=\mathbf{pa}) \\
        \end{aligned}
    \end{equation*}
    Now, we get,
    \begin{equation*}
        \begin{aligned}
            P(y \mid do(A=a),do(\mathbf{X=x}))=P(y \mid A=a,Pa'(Y)=\mathbf{pa})
        \end{aligned}
    \end{equation*}
\end{proof}

Theorem~\ref{theo:main} removes the descendant nodes of $Y$ from the conditioning set in the conditional probabilities for CDE estimation, and this removes possible collider bias. Furthermore, it gives a succinct set of variables for estimating CDE.

\begin{figure}[t]
	\centering
	\includegraphics[width=0.7\linewidth]{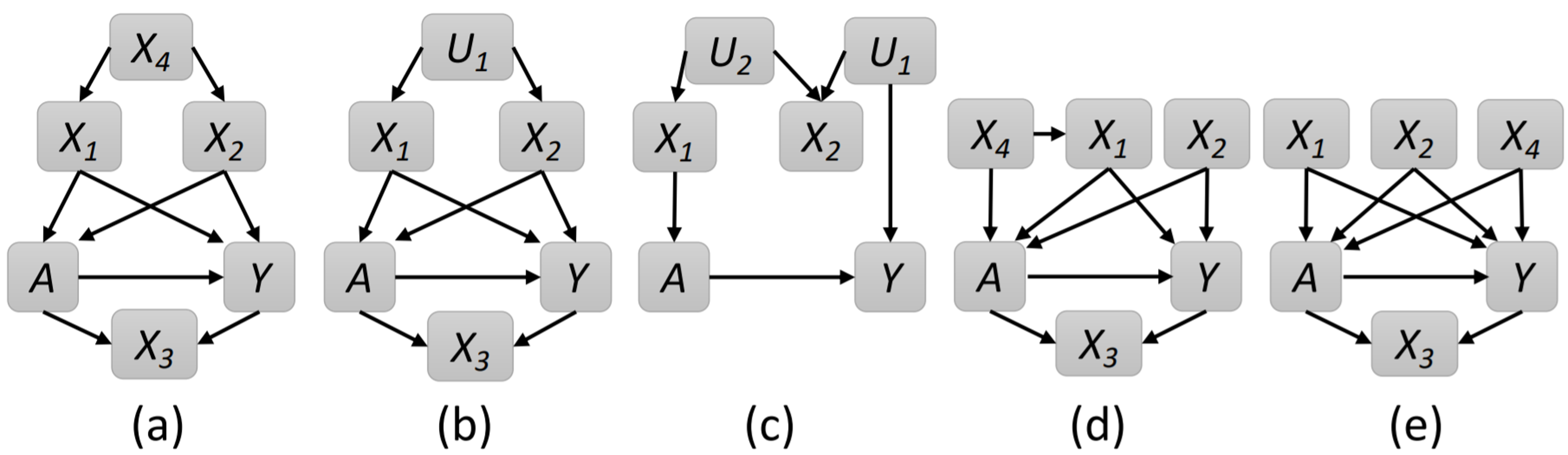}
	\vspace{-0.2cm}
	\caption{DAGs for the examples of Theorem/Corollary. $X_1$ to $X_6$ are observed variables, and $U_1$ and $U_2$ are unobserved variables.}
	\label{pic:TheoremEx}
	\vspace{-0.5cm}
\end{figure}

For example, in Figure~\ref{pic:TheoremEx}(a), $P(y \mid do(a), do(x_1, x_2, x_3, x_4)) = P(y \mid a, x_1, x_2)$ based on Theorem~\ref{theo:main}, where we use $x_i$ for $X_i = x_i$. The CDE is determined by conditional probabilities on $A$, $X_1$ and $X_2$. Since $X_3$ is not used in the conditioning set, there will be no collider bias. Theorem~\ref{theo:main} is based on the causal sufficiency assumption, which assumes that there are no unobserved common causes in the data set. In real-world applications, unobserved variables are unavoidable. When there are unobserved variables, how do we estimate CDE?Let us consider two types of unobserved variables. 
The first type of unobserved common causes are in between a parent node and $Y$. In this case, an edge emitting from a unobserved common cause goes into the parent node, and the parent node may form a collider with another incoming edge from another observed or unobserved variable (see the following for an example in Figure~\ref{pic:TheoremEx} (c)). This will invalidate the result of Theorem~\ref{theo:main}. The following Corollary will fix this. The second type of unobserved common causes are between other nodes apart from the parent nodes and $Y$. The proof in the following corollary will show that they do not invalidate the result of Theorem~\ref{theo:main}.           

\begin{corollary}
    \label{theo:allcauses}
    Let $Ca(Y)$ include all the direct causes and only direct causes of $Y$ except $A$. $P(y \mid do(A=a), do(\mathbf{X=x})) = P(y \mid A=a, Ca(Y) = \mathbf{ca})$.
\end{corollary}

\begin{proof}

Direct causes of $Y$ will be parent nodes of $Y$ in any DAG even when the unobserved common causes are included, i.e. the causal sufficiency is unsatisfied. Since $Pa'(Y) = Ca(Y)$ and there is not an unobserved variable in between a direct cause and $Y$, $P(y \mid do(A=a), do(\mathbf{X=x})) = P(y \mid A=a, Ca(Y) = \mathbf{ca})$ can be derived following the same procedure in Theorem~\ref{theo:main}.

Since other variables apart from $Pa'(Y)$ are not used in reducing $P(y \mid do(A=a), do(\mathbf{X=x}))$, the unobserved common casues between these variables are irrelevant to the deduction and do not affect the above conclusion. 
%
%
%
%
\end{proof}

Corollary~\ref{theo:allcauses} indicates that discrimination detection is sound when the audit agency knows all the direct causes of $Y$ and uses them as the conditioning set when calculating CDE. For example, in Figure~\ref{pic:TheoremEx}(b), $P(y \mid do(a), do(x_1, x_2, x_3, U_1))\\=P(y \mid a,x_1,x_2)$ based on Corollary~\ref{theo:allcauses}. Unobserved ancestral variables of $Y$ are blocked off from $Y$ by $X_1$ and $X_2$, and they do not affect the probability of $Y$. The unobservable variables can be in the descendant nodes of $Y$ too, but they do not affect the CDE estimation since they will not be used anyway.

We will further explain why direct causes are necessary for Corollary~\ref{theo:allcauses}. Let Figure~\ref{pic:TheoremEx}(c) be a true DAG with two unobserved variables $U_1$ and $U_2$. $X_2$ is not a direct cause of $Y$. Since $U_1$ and $U_2$ are unobserved, $X_2$ is perceived as a parent of $Y$ in the observed data. If $X_2$ is used to estimate CDE, the estimation will be biased since the back door path ($Y, U_1, U_2, X_1, A$) is opened when $X_2$ is conditioned on. In this case, $X_1$ is necessary to block the path. When both $X_1$ and $X_2$ are included, the CDE estimation is unbiased. Sometimes, we need redundancy to prevent such a biased estimation.  

Both Theorem~\ref{theo:main} and Corollary~\ref{theo:allcauses} give a succinct conditioning set for CDE estimation. In fact, a superset of the direct causes works as long as the superset does not contain descendant nodes of $Y$. In a DAG, ancestral nodes represent the direct causes and indirect causes of $Y$.  

\begin{corollary}
    \label{theo:antecedent}
    Let $\mathbf{B}$ include all the direct causes and some (or all) indirect causes of $Y$. We have $P(y \mid do(A=a),do(\mathbf{X=x})) = P(y \mid A=a,  \mathbf{B=b})$.
\end{corollary}

\begin{proof}
	 Let $\mathbf{B} = Ca'(Y) \cup \mathbf{R}$, and $\mathbf{R}$ includes indirect causes of $Y$. Following Corollary~\ref{theo:allcauses}, $P(y \mid do(A=a), do(\mathbf{X=x})) = P(y \mid A=a,  Ca'(Y) = \mathbf{ca})$. Based on the Markov condition~\ref{Markovcondition}, $Y$ is independent of $\mathbf{R}$ given $A \cup Ca'(Y)$. Hence,  $P(y \mid A=a,  Ca'(Y) = \mathbf{ca}) = P(y \mid A=a, \mathbf{B} )$. 
\end{proof}

Corollary~\ref{theo:antecedent} allows some redundancy in the conditioning set  comparing to Corollary~\ref{theo:allcauses}. In practice, the redundancy gives a flexibility for users to determine the direct causes of $Y$. Sometimes, a direct cause and an indirect cause are difficult to distinguish, and Corollary~\ref{theo:antecedent} indicates that including both does not bias the CDE estimation. For example, in Figure~\ref{pic:TheoremEx}(d), $P(y \mid do(a), do(x_1, x_2, x_3, x_4))\\=P(y \mid a, x_1, x_2)=P(y \mid a, x_1, x_2, x_4)$ based on Corollary~\ref{theo:antecedent} if $X_1$ and $X_2$ are all direct causes of $Y$. Let us assume that Figure~\ref{pic:TheoremEx}(d) is the true DAG, but a government agency has a DAG as Figure~\ref{pic:TheoremEx}(e) since they do not know which one of $X_1 $ and $X_4$ is the direct cause of $Y$. A CDE estimation based on the imprecise DAG in Figure~\ref{pic:TheoremEx}(e), i.e.  $P(y\mid do(a), do(x_1, x_2, x_3, x_4))=P(y \mid a, x_1, x_2, x_4)$ is also unbiased. 

\section{Implementing Unbiased Situation Test}
\label{sec:algo}

We summarise the discussion and propose the following unbiased situation test.

\begin{definition}[Unbiased Situation Test (UST)]
    \label{def_UST}
    UST exams whether a classifier $\hat{Y} = f()$ is fair for a given case $\mathbf{x_i}$ by calculating $CDE(\mathbf{x_i}) = P(\hat{Y} = 1 \mid  A=1, \mathbf{B = b_i}) - P(\hat{Y} = 1 \mid A=0, \mathbf{B = b_i})$, where $\mathbf{B}$ is the set of direct causes and some (or all) indirect causes of $Y$. The test case $\mathbf{x}_i$ is discriminated if $|CDE(\mathbf{x_i})| \ge \tau$, where $\tau$ is a threshold specified by the regulatory policy.
\end{definition}

All variables in the problem except $(A, Y)$ can be categorized into two types: $\mathbf{B}$ and $\mathbf{C}$. $\mathbf{B}$ is the set of ancestral nodes of $Y$ which can be identified by the regulatory policy, and $\mathbf{C}$ includes others. Note that irrelevant variables which are independent of $Y$ are not in $\mathbf{X}$.

\begin{algorithm}[t]
	\caption{Unbiased Situation Test (UST)}
	\label{pseudocode01}
	\small
	\begin{flushleft}
		\noindent {\textbf{Input}}: Classifier $f()$, $\mathbf{X= B \cup C}$ as defined in the text. $P(\mathbf{C = c_i})$. Test cases $D_{Test}$. The threshold $\tau$.\\
		\noindent {\textbf{Output}}: $L$, a list of discriminated cases in $D_{Test}$.\\
	\end{flushleft}
	\begin{algorithmic}[1]
		\vspace{-0.4cm}
		\FOR{each $r_i \in D_{Test}$}
		\STATE {Let $r_i'$ be the record by flipping the value of $A$ in $r_i$}
		\STATE {Let  $P(\hat{Y} =1  \mid r_i) = f(r_i)$ and $P(\hat{Y} =1 \mid r'_i) = f(r_i')$}
		\STATE {Obtain $ P(\hat{Y} =1 \mid A=A(r_i), \mathbf{B} = B(r_i))$ and $P(\hat{Y} =1  \mid  A=A(r'_i), \mathbf{B}=B(r'_i))$ by Equation~\ref{marginalisation} where $A()$ and $B()$ return values of $A$ and $\mathbf{B}$ in the records respectively}
		\STATE {Conduct situation test by Definition \ref{def_UST} and update $L$}
		\ENDFOR
		\STATE {Return $L$}
	\end{algorithmic}
\end{algorithm}

To conduct UST as in Definition \ref{def_UST}, one problem is that an audit agency cannot obtain the conditional probability $P(\hat{Y} =1 \mid A = a, \mathbf{B} = \mathbf{b}_{i})$ directly since it does not access the training data set or a unbiased sample of the population. Instead, it can have $P(\hat{Y} =1 \mid A = a, \mathbf{B = b_i, C = c_i})$ from the classifier $f()$. Therefore, the following marginalisation is used:

\vspace{-0.2cm}
\begin{equation}
    \label{marginalisation}
    \begin{aligned}
    P(\hat{y} \mid A = a, \mathbf{B= b_i}) = \sum_{\mathbf{c_i} \in \mathbf{C}} P(\hat{y}  \mid  a, \mathbf{b_i, C = c_i}) P(\mathbf{c_i}) 
    \end{aligned}
\vspace{-0.1cm}
\end{equation}
where $\hat{y}$ denotes $\hat{Y} = 1$, and probability $P(\mathbf{c_i})$ can be obtained from some sources, such as government census data. The algorithm for UST is presented in Algorithm~\ref{pseudocode01}. 

The complexity for UST algorithm is $O(n)$, where $n$ is the size of $D_{Test}$, i.e. linear to the number of test records. 

\section{Experiments}

In this section, we first demonstrate UST algorithm can correct the spurious associations generated by collider. Then, we simulate the audit process by using real-world data set. We only compare UST with NST in population-level sampling since other situation test methods, such as, CST~\cite{zhang2016situation}, k-NN based situation test~\cite{luong2011k} need to access training data set which is unavailable in our problem. We also demonstrate that data-based audit method fails in unrepresentative sampling but UST works. Finally, we apply the UST algorithm to compare fairness for different models and guide the audit agency to choose the model. The experimental settings and details can be found in the Appendix.

\subsection{Correcting collider biases} 

We construct synthetic data sets including a collider as discussed in the Appendix. UST has significantly reduced biases in CDE estimation. Bias is used to measure the error between an estimated CDE and the true CDE. Biases of including and not including a collider are shown in Table~\ref{tab:RESsynthetic}. The later is the UST method which corrects biases of a collider in data by directly using Corollary~\ref{theo:antecedent}. 

\vspace{-0.5cm}
\begin{table}
	\begin{minipage}[t]{0.5\linewidth}
		\centering
		\caption{UST has significantly reduced biases caused by a collider.}
		\vspace{-0.2cm}
		\label{tab:RESsynthetic}
		{\scriptsize \begin{tabular}{ccc}
				\specialrule{0.1em}{0pt}{0.3pt}
				Trials & Bias (with collider) & Bias (UST) \\ 
				\specialrule{0.05em}{-1pt}{0.3pt}
				1 & 0.143 $\pm$ 0.011 & 0.072 $\pm$ 0.004 \\
				2 & 0.154 $\pm$ 0.013 & 0.074 $\pm$ 0.004 \\
				3 & 0.149 $\pm$ 0.012 & 0.066 $\pm$ 0.004 \\
				4 & 0.149 $\pm$ 0.014 & 0.067 $\pm$ 0.004 \\
				5 & 0.152 $\pm$ 0.012 & 0.069 $\pm$ 0.004 \\ 
				\specialrule{0.1em}{-1pt}{0.3pt}
		\end{tabular}}
	\end{minipage}\quad
	\begin{minipage}[t]{0.5\linewidth}
		\centering
		\caption{Suburb variable (collier) improves the accuracy of classification models.}
		\vspace{-0.2cm}
		\label{tab:ACC}
		{\scriptsize \begin{tabular}{ccc}
				\specialrule{0.1em}{0pt}{0.3pt}
				& Acc. w/ Sub & Acc. w/o Sub \\ 
				\specialrule{0.05em}{-1pt}{0.3pt}
				DT & 89.66\% & 81.01\% \\
				SVM & 89.60\% & 80.93\% \\
				RF & 89.81\% & 80.86\% \\
				NN & 89.64\% & 80.85\% \\ 
				\specialrule{0.1em}{-1pt}{0.3pt}
		\end{tabular}}
	\end{minipage}
\end{table}
\vspace{-1cm}

\subsection{Simulating an audit process using Adult data set}

\begin{wrapfigure}{r}{0.4\textwidth}
	\vspace{-1.1cm}
	\begin{center}
		\includegraphics[width=\linewidth]{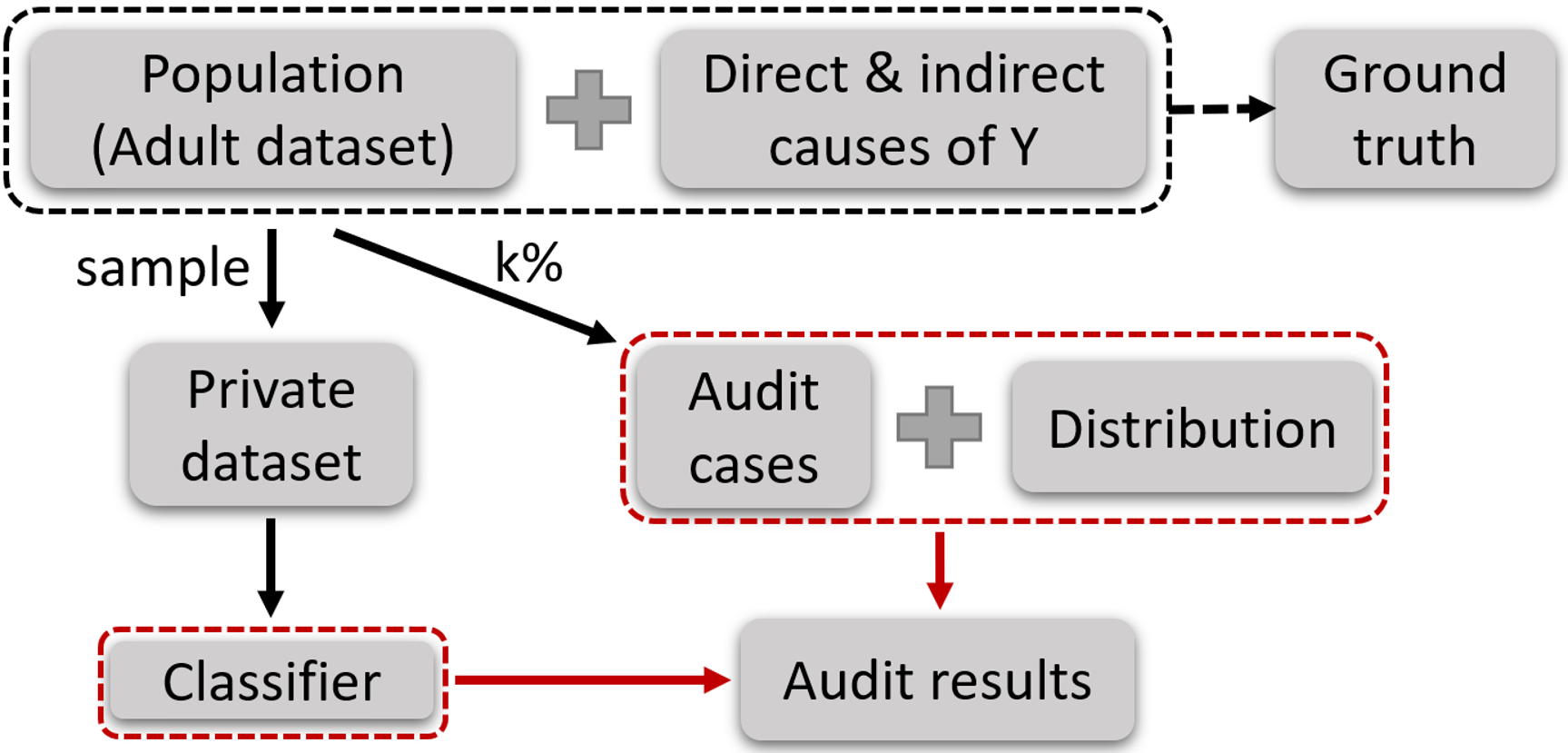}
	\end{center}
	\vspace{-0.5cm}
	\caption{A simulation of audit process using Adult data set}
	\label{pic:Adult_process}
	\vspace{-0.8cm}
\end{wrapfigure}

The Adult data set from UCI Machine Learning Repository \cite{adult1998data} is used to simulate audit process as shown in Figure~\ref{pic:Adult_process}.  We use the Adult data set as the population for generating the ground truths. A company has a sample (50\%) as the private data to build a model. The red dashed line represents the information that the audit agency has access to. The ground truths are generated from the population and all causes of $Y$.


Race is the protected variable and Salary is the outcome. Other variables are Education$\_$level, Marriage$\_$statues, Work$\_$hour, Work$\_$class, and they all determine the salary. We simulate a Suburb variable as a collider. The accuracy of a classifier is significantly higher when the Suburb is used than not as shown in Table \ref{tab:ACC}. The accuracy improvement is due to the spurious associations.

\subsection{Comparing the audit performance of NST and UST}


We apply UST to audit a few well-known classifiers built from sample data set. NST (introduced in the introduction) is used for the comparison since it is the only method for assessing the fairness of a classifier without accessing the training data set. 

We use precision and recall for the comparison. The ground truth for each audit case is calculated by using the population and the causes of $Y$. Audit cases are $k\%$ records randomly selected from the population. For each $k$, we resample audit cases 10 times and report the average precision and recall.

\begin{table*}[t]
	\centering
	\caption{The audit performance comparison of NST and UST. The higher values are highlighted. The standard error is shown in brackets.}
	\vspace{-0.2cm}
	\label{tab:REScasestudy}
	{\scriptsize \begin{tabular}{cccccccc}
			\specialrule{0.1em}{0pt}{0.3pt}
			&  & \multicolumn{2}{c}{k = 0.1\%} & \multicolumn{2}{c}{k = 0.5\%} & \multicolumn{2}{c}{k = 1\%} \vspace{-0.05cm}\\ \cline{3-8}
			&  & NST & UST & NST & UST & NST & UST \\ \specialrule{0.05em}{-2pt}{1pt}
			\multirow{2}{*}{DT} & Recall & 59.6\%(0.96) & \textbf{79.8\%(0.11)} & 56.7\%(0.27) & \textbf{73.3\%(0.05)} & 56.3\%(0.05) & \textbf{71.7\%(0.10)} \\
			& Precision & 84.4\%(0.32) & \textbf{98.1\%(0.16)} & 78.9\%(0.06) & \textbf{99.1\%(0.01)} & 80.3\%(0.02) & \textbf{98.8\%(0.01)} \\ \specialrule{0.05em}{-1pt}{1pt}
			\multirow{2}{*}{SVM} & Recall & 77.9\%(1.18) & \textbf{87.8\%(0.26)} & 75.7\%(0.17) & \textbf{89.9\%(0.03)} & 74.1\%(0.06) & \textbf{89.1\%(0.02)} \\
			& Precision & 68.1\%(0.73) & \textbf{83.4\%(0.14)} & 65.1\%(0.11) & \textbf{79.9\%(0.06)} & 64.6\%(0.13) & \textbf{81.0\%(0.05)} \\ \specialrule{0.05em}{-1pt}{1pt}
			\multirow{2}{*}{RF} & Recall & 56.1\%(1.91) & \textbf{73.8\%(0.32)} & 58.2\%(0.23) & \textbf{66.8\%(0.03)} & 57.7\%(0.09) & \textbf{65.2\%(0.14)} \\
			& Precision & 88.6\%(0.17) & \textbf{96.4\%(0.25)} & 86.8\%(0.02) & \textbf{97.8\%(0.01)} & 86.9\%(0.04) & \textbf{98.4\%(0.01)} \\ \specialrule{0.05em}{-1pt}{1pt}
			\multirow{2}{*}{NN} & Recall & 65.9\%(1.17) & \textbf{74.9\%(0.14)} & 67.6\%(0.16) & \textbf{71.5\%(0.10)} & 67.9\%(0.06) & \textbf{69.2\%(0.08)} \\
			& Precision & 85.9\%(0.24) & \textbf{96.7\%(0.21)} & 81.7\%(0.04) & \textbf{97.1\%(0.01)} & 82.8\%(0.02) & \textbf{97.2\%(0.01)} \\
			\specialrule{0.1em}{-1pt}{1pt}
	\end{tabular}}
\end{table*}

UST outperforms NST in both precision and recall as shown in Table \ref{tab:REScasestudy}. With the increasing number of audit cases, the deviations of both methods decrease. From the gaps between the precision and recall of NST and UST, we see that the collider bias deteriorates the detection performance of NST significantly.

\subsection{Data based audit may be biased} 
Removing the collider from the data can be used as an alternative method to UST. However, a data-based audit (DBA) relies on the representativeness of the audit cases for the population. The representativeness is difficult to be ensured because individuals who receive unfair treatments likely require the authority to audit their results. An audit agency does not have a resource to collect a representative sample for auditing.


We simulate the unrepresentative audit cases by over (under) sampling discriminatory cases in the population. In the Adult data set, about 15\% of the individuals are discriminatory and this ratio is the baseline. We vary discriminatory ratios in $k=1\%$ audit cases.

\begin{table*}[t]
	\centering
	\caption{The audit performance comparison of DBA and UST with discriminatory sample. The higher values are highlighted. The standard error is shown in brackets.}
	\vspace{-0.2cm}
	\label{tab:BDM}
	{\scriptsize \begin{tabular}{cccccccc}
			\specialrule{0.1em}{0pt}{0.3pt}
			&  & \multicolumn{2}{c}{Discriminatory Ratio=0\%} & \multicolumn{2}{c}{Discriminatory Ratio=10\%} & \multicolumn{2}{c}{Discriminatory Ratio=20\%} \\  \cline{3-8}
			&  & DBA & UST & DBA & UST & DBA & UST \\ \specialrule{0.05em}{-2pt}{1pt}
			\multirow{2}{*}{DT} & Recall & 60.1\%(1.18) & \textbf{72.2\%(0.06)} & 57.8\%(1.72) & \textbf{71.7\%(0.07)} & 60.9\%(0.91) & \textbf{71.8\%(0.10)} \\
			& Precision & 84.6\%(0.23) & \textbf{97.5\%(0.01)} & 78.2\%(0.35) & \textbf{96.2\%*(0.01)} & 70.9\%(0.11) & \textbf{95.1\%(0.01)} \\ \specialrule{0.05em}{-1pt}{1pt}
			\multirow{2}{*}{SVM} & Recall & 39.8\%(1.64) & \textbf{89.0\%(0.01)} & 40.0\%(1.51) & \textbf{89.5\%(0.01)} & 35.6\%(2.57) & \textbf{89.8\%(0.01)} \\
			& Precision & 77.8\%(0.19) & \textbf{82.5\%(0.03)} & 71.2\%(0.31) & \textbf{76.0\%(0.05)} & 56.4\%(1.02) & \textbf{67.7\%(0.06)} \\ \specialrule{0.05em}{-1pt}{1pt}
			\multirow{2}{*}{RF} & Recall & 39.9\%(0.18) & \textbf{65.6\%(0.06)} & 40.4\%(0.14) & \textbf{65.2\%(0.06)} & 39.5\%(1.13) & \textbf{65.2\%(0.05)} \\
			& Precision & 78.9\%(0.14) & \textbf{97.3\%(0.02)} & 72.6\%(0.16) & \textbf{96.4\%(0.02)} & 61.2\%(0.26) & \textbf{94.2\%(0.05)} \\ \specialrule{0.05em}{-1pt}{1pt}
			\multirow{2}{*}{NN} & Recall & 46.1\%(2.02) & \textbf{69.8\%(0.07)} & 45.7\%(2.11) & \textbf{69.8\%(0.06)} & 39.4\%(2.26) & \textbf{69.4\%(0.09)} \\
			& Precision & 80.6\%(0.11) & \textbf{95.0\%(0.03)} & 73.6\%(0.29) & \textbf{93.0\%(0.03)} & 60.0\%(0.51) & \textbf{90.1\%(0.07)} \\ 
			\specialrule{0.05em}{-1pt}{1pt}
	\end{tabular}}
	\vspace{-0.5cm}
\end{table*}

The performance of DBA deteriorates significantly when a discriminatory ratio deviates from the baseline as shown in Table \ref{tab:BDM}. Note that all discrimination detection methods based on data have the same limitation. In contrast, UST maintains similar performance.

\subsection{Rank models based on fairness}
We show that UST can be used for comparing the fairness of different models. We first discuss the metrics for the comparison. After discrimination detection on a model using audit cases with ground truths, we obtain True Positive (TF), False Positive (FP), True Negative (TN), and False Negative (FN).  

FN indicates the cases that are unfair but are corrected to be fair by the model. They are favourable for fair predictions, and we use correction rate, $CR = \frac{FN}{TP+FN}$, to represent the proportion of true unfair cases being corrected by a model. In contrast, FP represents that the cases that are fair become unfair after model predictions. These cases are called reversed discrimination and are unfavourite for predictions. We use the reversion rate, $RR = \frac{FP}{FP+TN}$, to represent the proportion of fair cases being reversed by a model. We wish the revision rate is as small as possible. 

\begin{wraptable}{r}{0.45\textwidth}
	\vspace{-1cm}
	\caption{The audit results of various models.}
	\label{tab:Model}
	\vspace{-0.2cm}
	\begin{center}
		{\scriptsize \begin{tabular}{ccr}
				\specialrule{0.1em}{0pt}{0.3pt}
				& CR($\uparrow$) & \multicolumn{1}{c}{RR($\downarrow$)} \\ \specialrule{0.05em}{-1pt}{0.5pt}
				DT & 29.09\% $\pm$ 0.079 & 1.69\% $\pm$ 0.001 \\
				SVM & 10.05\% $\pm$ 0.024 & 39.37\% $\pm$ 0.049 \\
				RF & 35.24\% $\pm$ 0.147 & 1.07\% $\pm$ 0.001 \\
				NN & 31.48\% $\pm$ 0.064 & 2.76\% $\pm$ 0.001 \\ \specialrule{0.1em}{-1pt}{0.3pt}
		\end{tabular}}
	\end{center}
	\vspace{-0.8cm}
\end{wraptable}

The correction rate and reversion rate of four classifiers are shown in Table~\ref{tab:Model}. Random Forest is the fairest model based on the two measures. Random Forest has corrected 35.24\% of unfair cases, and only reversed 1.07\% of fair cases to unfair. Note that, their prediction accuracies are very similar, but their correction rate and reversion rate are different. This assessment shows that some errors made by a model are better than others in terms of the fairness of a model.

\section{Related Works}
The work belongs to discrimination detection. Detection methods are divided into the group, and individual-based. Another division of the methods is association or causal based. 

At the group level, a number of metrics have been defined to detect discrimination. Demographic parity, a well-known fairness measurement, is defined by \cite{dwork2012fairness,zemel2013learning}. Other measurements including equalised odds \cite{hardt2016equality}, predictive rate parity \cite{zafar2017fairness}. However, these group-based fairness does not necessarily mean individual fairness. Many algorithms focus on detecting discrimination at the individual level. Authors in~\cite{speicher2018unified} use existing inequality indices from economics to measure individual level fairness. Speicher et al.~\cite{lohia2019bias} propose an individual level discrimination detector, which is used to prioritise data samples and aims to improve the subgroup fairness measure of disparate impact. 

Under the causal framework, Li et al. \cite{li2017discrimination} use the (conditional) average causal effect to quantify fairness for (sub)group level discrimination detection.  Counterfactual fairness~\cite{kusner2017counterfactual} is an attractive definition of individual level fairness measurements by causality. It means that a decision is fair towards an individual if it is the same in both the actual world and a counterfactual world (when a value of a protected variable is changed). However, it needs strong assumptions.   authors in~ \cite{nabi2018fair,zhang2018fairness,zhang2017causal} use nature direct effect and nature indirect effect to quantify fairness. The path-specific causal effect \cite{chiappa2019path,wu2019pc,zhang2018causal} have been used to quantify fairness when the regulatory policy recognises some causal paths involving a protected variable fair. Nature direct (indirect) effect and path-specific causal effect all need counterfactual reasoning and are difficult to implement in practice since the strong assumptions are related to counterfactual reasoning.  

Situation test related work has been discussed in the introduction and hence is omitted here. The above-mentioned related work only introduces some influential contributions. For more related work,  please refer to the literature review \cite{caton2020fairness,corbett2018measure,mehrabi2021survey,zhang2017anti}.

\section{Conclusions}
\label{sec:conclude}
In this paper, we have discussed collider bias in fairness assessment. We have presented theoretical results based on the graphical causal model to avoid collider biases in fairness assessment. The results are useful for discrimination detection and also for feature selection for building fair classifiers. We have proposed an Unbiased Situation Test (UST) algorithm for the fairness assessment of a classifier without accessing the training data set or a sample of the population. Experimental results show that UST effectively reduces collider biases and can be used to assess the fairness of a classifier without accessing to data. The UST is promising for an audit agency to audit machine learning models by private companies and organisations.

\subsubsection{Acknowledgements} We acknowledge Australian Research Council (DP200101210) and Natural Sciences and Engineering Research Council of Canada.

%
%
%
\bibliographystyle{splncs04}
\bibliography{mybibliography.bib}

\begin{thebibliography}{10}
\providecommand{\url}[1]{\texttt{#1}}
\providecommand{\urlprefix}{URL }
\providecommand{\doi}[1]{https://doi.org/#1}

\bibitem{bendick2007situation}
Bendick, M.: Situation testing for employment discrimination in the united
  states of america. Horizons strat{\'e}giques  \textbf{3},  17--39 (2007)

\bibitem{adult1998data}
Blake, C., Merz, C.: Uci machine learning repository.
  \url{https://archive.ics.uci.edu/ml/datasets/adult} (1998), accessed:
  1996-05-01

\bibitem{caton2020fairness}
Caton, S., Haas, C.: Fairness in machine learning: A survey. arXiv preprint
  arXiv:2010.04053  (2020)

\bibitem{chiappa2019path}
Chiappa, S.: Path-specific counterfactual fairness. In: AAAI. pp. 7801--7808
  (2019)

\bibitem{ConditioningOnCollider}
Cole, S.R., Platt, R.W., Schisterman, E.F., Chu, H., Westreich, D., Richardson,
  D., Poole, C.: Illustrating bias due to conditioning on a collider.
  International Journal of Epidemiology  \textbf{39(2)},  417--20 (2010)

\bibitem{corbett2018measure}
Corbett-Davies, S., Goel, S.: The measure and mismeasure of fairness: A
  critical review of fair machine learning  (2018)

\bibitem{dwork2012fairness}
Dwork, C., Hardt, M., Pitassi, T., Reingold, O., Zemel, R.: Fairness through
  awareness. In: ITCS. pp. 214--226 (2012)

\bibitem{feldman2015certifying}
Feldman, M., Friedler, S.A., Moeller, J., Scheidegger, C., Venkatasubramanian,
  S.: Certifying and removing disparate impact. In: KDD. pp. 259--268 (2015)

\bibitem{haggstrom2018data}
H{\"a}ggstr{\"o}m, J.: Data-driven confounder selection via markov and bayesian
  networks. Biometrics (2),  389--398 (2018)

\bibitem{hardt2016equality}
Hardt, M., Price, E., Srebro, N.: Equality of opportunity in supervised
  learning. NeurIPS pp. 3315--3323 (2016)

\bibitem{hernan2020causal}
Hern{\'a}n, M.A., Robins, J.M.: Causal Inference: What If. Boca Raton: Chapman
  \& Hall/CRC (2020)

\bibitem{kilbertus2017avoiding}
Kilbertus, N., Rojas-Carulla, M., Parascandolo, G., Hardt, M., Janzing, D.,
  Sch{\"o}lkopf, B.: Avoiding discrimination through causal reasoning. In:
  NeurIPS. pp. 656--666 (2017)

\bibitem{kusner2017counterfactual}
Kusner, M., Loftus, J., Russell, C., Silva, R.: Counterfactual fairness. In:
  NeurIPS. pp. 4069--4079 (2017)

\bibitem{li2017discrimination}
Li, J., Liu, J., Liu, L., Le, T.D., Ma, S., Han, Y.: Discrimination detection
  by causal effect estimation. In: BigData. pp. 1087--1094. IEEE (2017)

\bibitem{lohia2019bias}
Lohia, P.K., Ramamurthy, K.N., Bhide, M., Saha, D., Varshney, K.R., Puri, R.:
  Bias mitigation post-processing for individual and group fairness. In:
  ICASSP. pp. 2847--2851. IEEE (2019)

\bibitem{luong2011k}
Luong, B.T., Ruggieri, S., Turini, F.: k-nn as an implementation of situation
  testing for discrimination discovery and prevention. In: KDD. pp. 502--510
  (2011)

\bibitem{mehrabi2021survey}
Mehrabi, N., Morstatter, F., Saxena, N., Lerman, K., Galstyan, A.: A survey on
  bias and fairness in machine learning. CSUR (6),  1--35 (2021)

\bibitem{nabi2018fair}
Nabi, R., Shpitser, I.: Fair inference on outcomes. In: AAAI (2018)

\bibitem{pearl2009causality}
Pearl, J.: Causality. Cambridge university press (2009)

\bibitem{PearlMackenzie18}
Pearl, J., Mackenzie, D.: The Book of Why. Basic Books, New York (2018)

\bibitem{ramsey2018tetrad}
Ramsey, J.D., Zhang, K., Glymour, M., Romero, R.S., Huang, B., Ebert-Uphoff,
  I., Samarasinghe, S., Barnes, E.A., Glymour, C.: Tetrad—a toolbox for
  causal discovery. In: 8th International Workshop on Climate Informatics
  (2018)

\bibitem{speicher2018unified}
Speicher, T., Heidari, H., Grgic-Hlaca, N., Gummadi, K.P., Singla, A., Weller,
  A., Zafar, M.B.: A unified approach to quantifying algorithmic unfairness:
  Measuring individual \&group unfairness via inequality indices. In: KDD. pp.
  2239--2248 (2018)

\bibitem{spirtes1991algorithm}
Spirtes, P., Glymour, C.: An algorithm for fast recovery of sparse causal
  graphs. Social science computer review pp. 62--72 (1991)

\bibitem{spirtes2000causation}
Spirtes, P., Glymour, C.N., Scheines, R., Heckerman, D.: Causation, prediction,
  and search. MIT press (2000)

\bibitem{wu2019pc}
Wu, Y., Zhang, L., Wu, X., Tong, H.: Pc-fairness: A unified framework for
  measuring causality-based fairness. NeurIPS  (2019)

\bibitem{zafar2017fairness}
Zafar, M.B., Valera, I., Gomez~Rodriguez, M., Gummadi, K.P.: Fairness beyond
  disparate treatment \& disparate impact: Learning classification without
  disparate mistreatment. In: WWW. pp. 1171--1180 (2017)

\bibitem{zemel2013learning}
Zemel, R., Wu, Y., Swersky, K., Pitassi, T., Dwork, C.: Learning fair
  representations. In: ICML. pp. 325--333 (2013)

\bibitem{zhang2018fairness}
Zhang, J., Bareinboim, E.: Fairness in decision-making—the causal explanation
  formula. In: AAAI (2018)

\bibitem{zhang2017anti}
Zhang, L., Wu, X.: Anti-discrimination learning: a causal modeling-based
  framework. IJDSA  \textbf{4}(1),  1--16 (2017)

\bibitem{zhang2016situation}
Zhang, L., Wu, Y., Wu, X.: Situation testing-based discrimination discovery: a
  causal inference approach. In: IJCAI. pp. 2718--2724 (2016)

\bibitem{zhang2017causal}
Zhang, L., Wu, Y., Wu, X.: A causal framework for discovering and removing
  direct and indirect discrimination. In: IJCAI. pp. 3929--3935 (2017)

\bibitem{zhang2018causal}
Zhang, L., Wu, Y., Wu, X.: Causal modeling-based discrimination discovery and
  removal: Criteria, bounds, and algorithms. TKDE (11),  2035--2050 (2018)

\end{thebibliography}

\newpage
\section*{Appendix}
\label{sec:appendix}

\subsection*{Synthetic data sets}
\begin{figure}
	\begin{minipage}[t]{0.5\linewidth}
		\centering
		\includegraphics[width=0.4\linewidth]{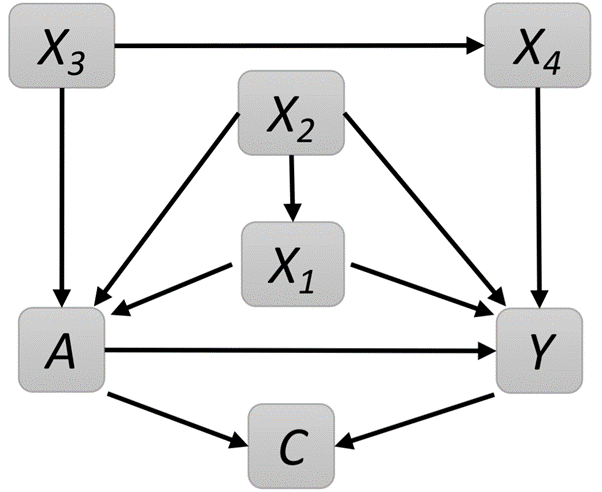}
		\caption{Causal relations in the synthetic data.}
		\label{pic:DAGsynthetic}
	\end{minipage}\quad
	\begin{minipage}[t]{0.5\linewidth}
		\centering
		\includegraphics[width=0.8\linewidth]{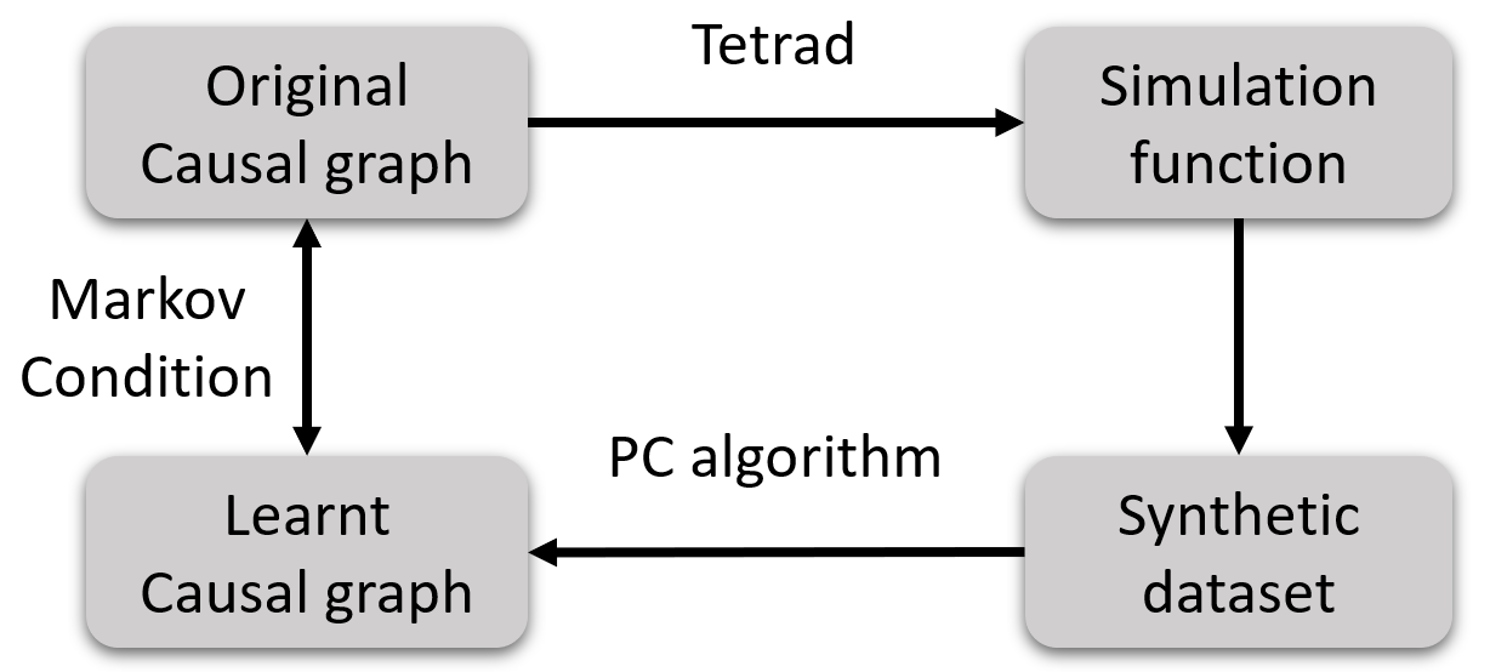}
		\caption{The process of generating synthetic data sets.}
		\label{pic:process}
	\end{minipage}
\end{figure}

We construct synthetic data sets by Tetrad \cite{ramsey2018tetrad}. We use a causal graph shown in Figure \ref{pic:DAGsynthetic} with $7$ variables, including a protected variable $A$, an outcome variable $Y$, other observed variables, $X_1,X_2,X_3,X_4$, and a collider variable $C$. 

We use the simulation function from Tetrad package to randomly generate $5$ data sets with 50,000 records. The use of 50,000 records is to ensure the faithfulness between the generated data and the causal graph. We employ the PC algorithm~\cite{spirtes1991algorithm,spirtes2000causation} to learn causal structure from synthetic data sets (setting parameter $\alpha = 0.05$). Every learnt causal graph is equivalent to the original causal graph under Markov Condition. The above process are shown in Figure \ref{pic:process}.

\vspace{-0.5cm}
\vspace{-1cm}

\subsection*{Adult data set}
\begin{wraptable}{r}{0.6\textwidth}
	\vspace{-1cm}
	\caption{The distribution for the variables in adult data set after preprocessing.}
	\label{tab:distribution}
	\vspace{-0.2cm}
	\begin{center}
		{\scriptsize \begin{tabular}{cllc}
				\specialrule{0.1em}{0pt}{0.3pt}
				Variable & \multicolumn{2}{c}{Value} & Distribution \\ \specialrule{0.05em}{-1pt}{0.3pt}
				\multirow{2}{*}{Race} & Black & 0 & 10.09\% \\
				& White & 1 & 89.91\% \\ \specialrule{0.05em}{-1pt}{0.3pt}
				\multirow{2}{*}{Education level} & Non-college & 0 & 45.74\% \\
				& College & 1 & 54.26\% \\ \specialrule{0.05em}{-1pt}{0.3pt}
				\multirow{3}{*}{Marriage statues} & Never-married & 0 & 32.83\% \\
				& Married & 1 & 47.12\% \\
				& Other statues & 2 & 20.05\% \\ \specialrule{0.05em}{-1pt}{0.3pt}
				\multirow{2}{*}{Work class} & Private company & 0 & 69.48\% \\
				& Other & 1 & 30.52\% \\ \specialrule{0.05em}{-1pt}{0.3pt}
				\multirow{2}{*}{Work hour} & Part-time & 0 & 22.29\% \\
				& Full-time & 1 & 77.71\% \\ \specialrule{0.05em}{-1pt}{0.3pt}
				\multirow{2}{*}{Salary} & Below 50k & 0 & 75.94\% \\
				& Above 50k & 1 & 24.06\% \\ 
				\specialrule{0.1em}{-1pt}{0.3pt}
		\end{tabular}}
	\end{center}
	\vspace{-0.8cm}
\end{wraptable}

The Adult data set from UCI Machine Learning Repository \cite{adult1998data} records 14 variables such as individual information, occupation information and education background on 48842 instances along with their salary. In this experiment, we only consider whether there is discrimination on race. We keep variables which are causes of salary, and remove records with non white/black, and with missing values. After data preprocessing, there are 46447 instances with 6 variables in the final data set. The distribution of the variables is shown in Table \ref{tab:distribution}.

\begin{figure}[t]
	\centering
	\includegraphics[width=0.6\linewidth]{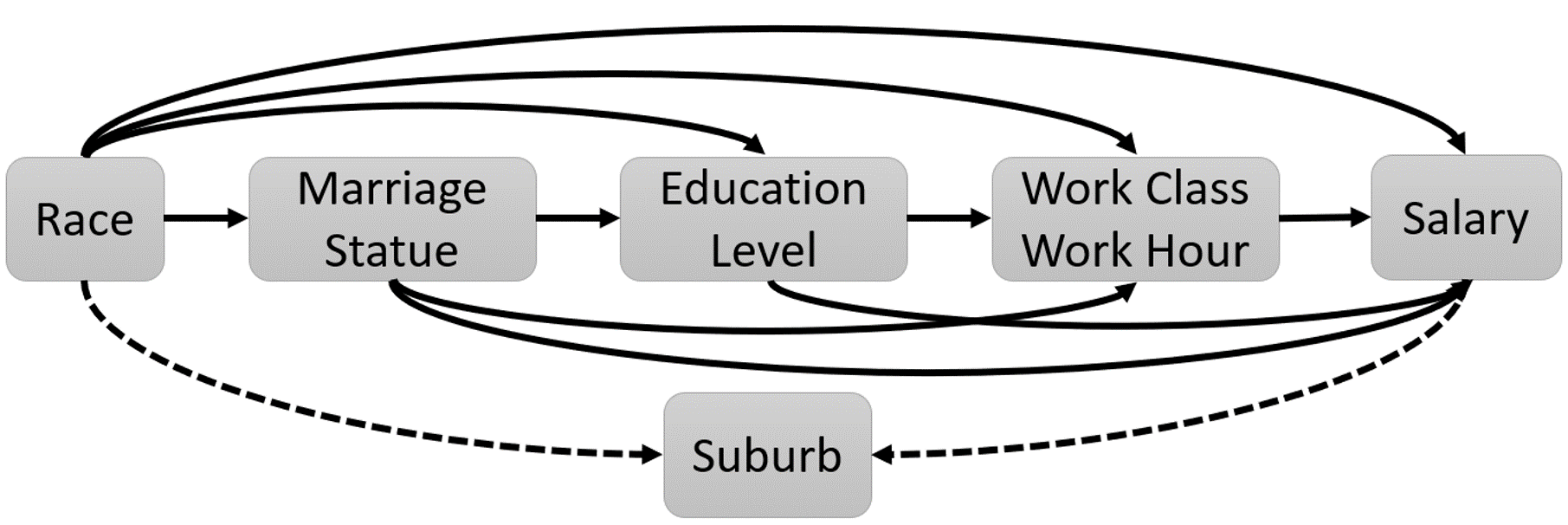}
	\caption{The causal graph for adult data set. The black lines show the original causal graph which is the same with \cite{nabi2018fair}. Work\_class and Work\_hour are shown as one variable to simplify the graph. The black dash lines link to the synthetic  collider.}
	\label{pic:adultdag}
\end{figure}

We use the same causal graph generated by \cite{nabi2018fair} in this experiment and assume it represents the regulations. Marriage$\_$statues, Education$\_$level, Work$\_$class, Work$\_$hour are legitimate variables to determine salary. The causal graph is drawn in Figure \ref{pic:adultdag}. In this causal graph, Work\_class and Work\_hour are shown in one variable since their relationships with other attributes are the same. 

We assume that suburb information is an effect of Salary and Race. The company makes use of Suburb to improve the performance of the classifier. The collider is represented by black dash line in Figure \ref{pic:adultdag}.

We follow the same procedure in \cite{haggstrom2018data} to generate the suburb for the adult data set. The generation rules are shown in Table \ref{tab:collider}. The distribution of collider is shown in Table \ref{tab:colliderdis}.

\vspace{-0.5cm}
\begin{table}
	\begin{minipage}[t]{0.5\linewidth}
		\centering
		\caption{The rules to generate collider variable for adult data set.}
		\vspace{-0.2cm}
		\label{tab:collider}
		{\scriptsize \begin{tabular}{ccc}
				\specialrule{0.1em}{0pt}{0.3pt}
				Race & Salary & Suburb \\ 
				\specialrule{0.05em}{-1pt}{0.3pt}
				black & below 50k & A \\
				white & below 50k & A \\
				black & above 50k & A \\
				white & above 50k & B \\ 
				\specialrule{0.1em}{-1pt}{0.3pt}
		\end{tabular}}
	\end{minipage}\quad
	\begin{minipage}[t]{0.5\linewidth}
		\centering
		\caption{The distribution for collider in adult data set.}
		\vspace{-0.2cm}
		\label{tab:colliderdis}
		{\scriptsize \begin{tabular}{cccc}
				\specialrule{0.1em}{0pt}{0.3pt}
				Variable & \multicolumn{2}{c}{Value} & Distribution \\ 
				\specialrule{0.05em}{-1pt}{0.3pt}
				\multirow{2}{*}{Suburb} & A & 0 & 63.10\% \\
				& B & 1 & 36.90\% \\ 
				\specialrule{0.1em}{-1pt}{0.3pt}
		\end{tabular}}
	\end{minipage}
\end{table}
\vspace{-0.5cm}

\end{document}